\pdfoutput=1

\documentclass[11pt]{article}

\usepackage{EMNLP2022}

\usepackage{times}
\usepackage{latexsym}

\usepackage[T1]{fontenc}

\usepackage[utf8]{inputenc}

\usepackage{microtype}

\usepackage{balance}
\usepackage{url}
\usepackage{graphicx}
\usepackage{amsmath,amssymb,amsfonts}
\usepackage{algorithmic}
\usepackage{textcomp}
\usepackage{xcolor}
\usepackage{tikz}
\usetikzlibrary{topaths,calc}
\usepackage[ruled]{algorithm2e}
\usepackage{subfigure}
\usepackage{booktabs}
\usepackage{balance}
\usepackage{colortbl}
\usepackage{enumitem}
\usepackage{multirow}
\usepackage{bbold}
\usepackage[normalem]{ulem}

\newcommand{\deer}{{\textbb{DEER}}}
\newcommand\deeremoji{\raisebox{-2pt}{\includegraphics[width=0.9em]{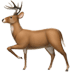}}}

\usepackage{etoolbox}

\usepackage{inconsolata}

\title{\deer{}\deeremoji{}: Descriptive Knowledge Graph for Explaining Entity Relationships}

\author{Jie Huang$^{*,1}$ $\quad$ Kerui Zhu$^{*,1}$ $\quad$ Kevin Chen-Chuan Chang$^{1}$ \\ \textbf{Jinjun Xiong$^{2}$ $\quad$ Wen-mei Hwu$^{1,3}$} \\
 $^1$University of Illinois at Urbana-Champaign, USA \\
 $^2$University at Buffalo, USA \\
 $^3$NVIDIA, USA \\
 \texttt{\{jeffhj, keruiz2, kcchang, w-hwu\}@illinois.edu} \\
 \texttt{jinjun@buffalo.edu}
}

\begin{document}
\maketitle
\begin{abstract}
We propose \deer{}\deeremoji{} (\textbf{D}escriptive Knowledge Graph for \textbf{E}xplaining \textbf{E}ntity \textbf{R}elationships) -- an open and informative form of modeling entity relationships. In \deer{}, relationships between entities are represented by free-text relation descriptions.
For instance, the relationship between entities of \textit{machine learning} and \textit{algorithm} can be represented as ``\emph{Machine learning} explores the study and construction of \emph{algorithms} that can learn from and make predictions on data.''
To construct \deer{}, we propose a self-supervised learning method to extract relation descriptions with the analysis of dependency patterns and generate relation descriptions with a transformer-based relation description synthesizing model, where no human labeling is required. Experiments demonstrate that our system can extract and generate high-quality relation descriptions for explaining entity relationships. The results suggest that we can build an open and informative knowledge graph without human annotation.\footnote{Code and data are available at \url{https://github.com/jeffhj/DEER}. $^*$Asterisk indicates equal contribution.}
\end{abstract}

\section{Introduction}

\begin{figure*}[tp!]
\centerline{\includegraphics[width=0.8\linewidth]{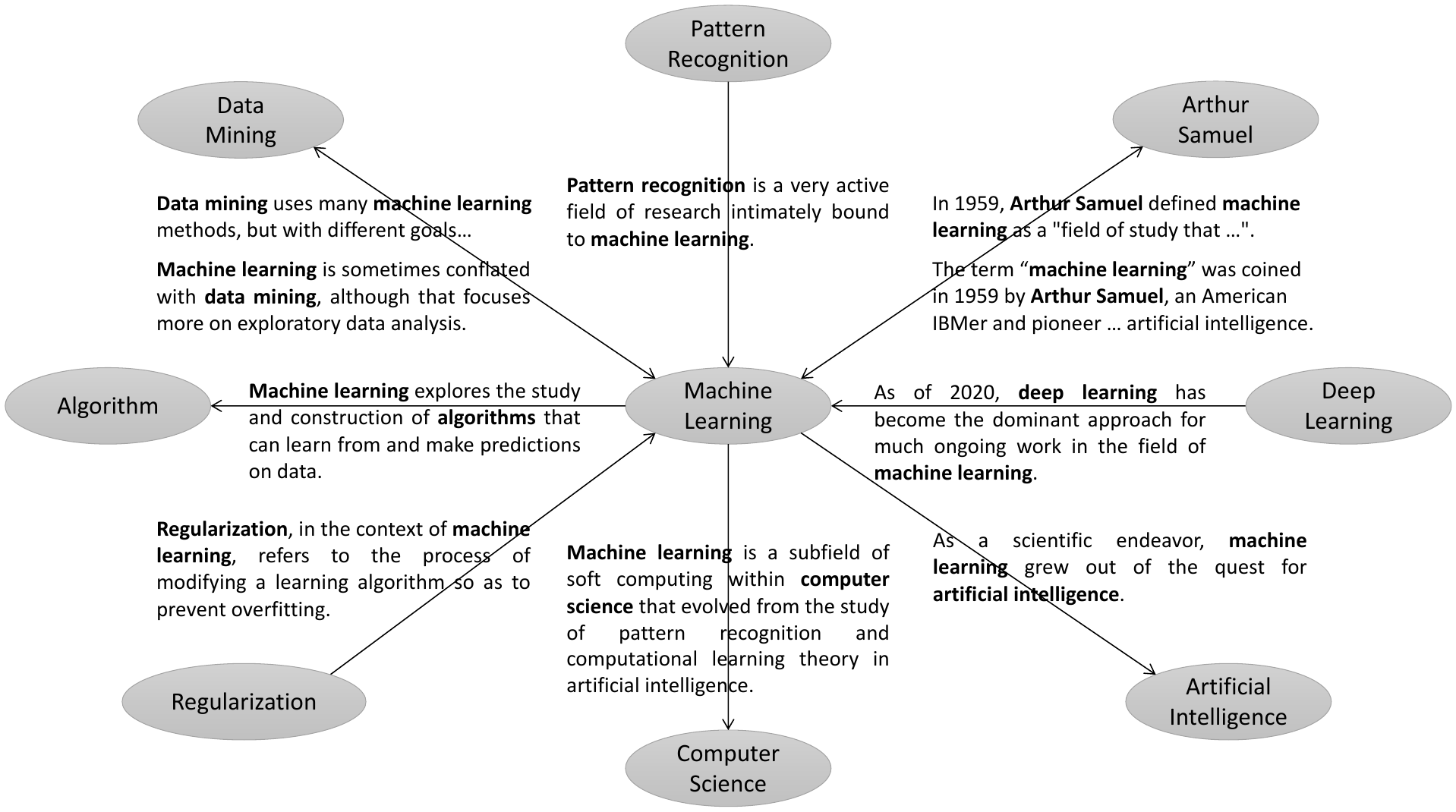}}
\vspace{-2mm}
\caption{Relations in \deer{}\deeremoji{}. Here we show \textit{machine learning} and several of its related entities, with corresponding relation descriptions produced by our model (only extraction) in the edges.}
\vspace{-5mm}
\label{fig:graph}
\end{figure*}

\begin{figure}[tp!]
\centerline{\includegraphics[width=0.86\linewidth]{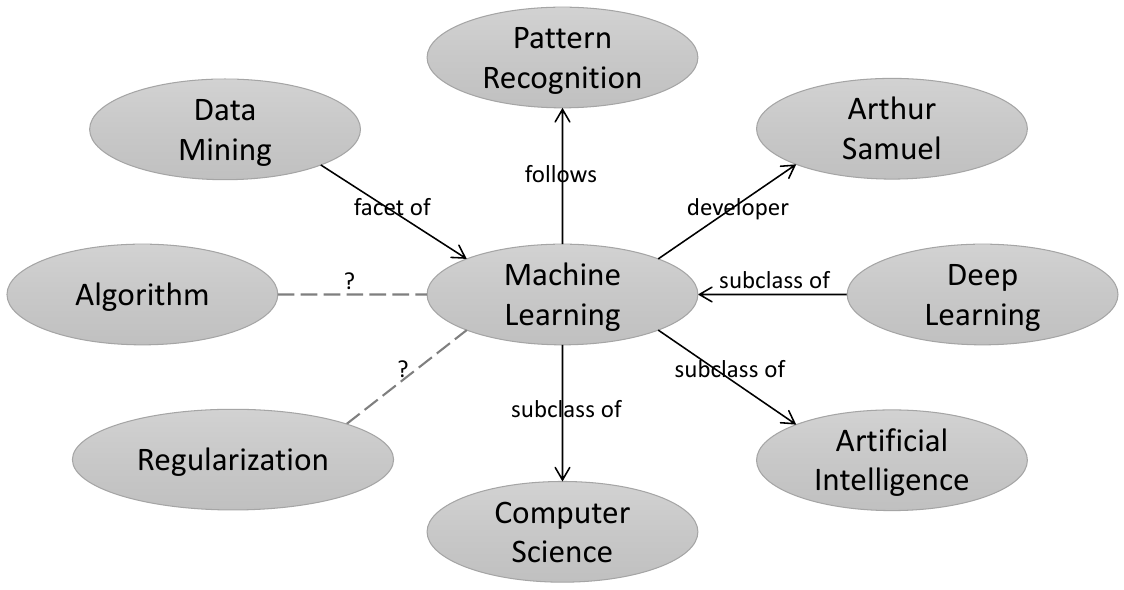}}
\vspace{-1.7mm}
\caption{Relations in Wikidata (Knowledge Graph), where ? means the relation is not present in the graph.}
\vspace{-5mm}
\label{fig:pkg}
\end{figure}

Relationships exist widely between entities. For example, a person may be related to another person or an institution, and a scientific concept can be connected to another concept.
At the same time, relationships between entities can be subtle or complex, e.g., the relationship between \textit{machine learning} and \textit{algorithm}.

To model relationships between entities, researchers usually construct knowledge graphs (KGs) \cite{ji2021survey,hogan2021knowledge}, where nodes are entities, e.g., \textit{machine learning}, and edges are relations, e.g., \textit{subclass of} (Figure \ref{fig:pkg}).
However, KGs usually require a pre-specified set of relation types, and the covered relation types are usually coarse-grained and simple. 
This indicates existing KGs lack two desired features. 
The first is \textit{\textbf{openness}}: for entities with a relationship not covered by the type set, KGs cannot handle their relationship directly. 
Besides, in many cases, the relationship between entities is complex or idiosyncratic that it cannot be simply categorized to a relation type. For instance, for related entities \textit{machine learning} and \textit{algorithm}, 
Wikidata \cite{vrandevcic2014wikidata}
does not include a relation for them, and it is also not easy to come up with a relation type to describe their relationship.  

The second feature is about \textit{\textbf{informativeness}}. With the relational facts in KGs, humans may still have difficulty in understanding entity relationships. For instance, from fact ``(data mining, \textit{facet of}, database)'' in Wikidata, humans may guess \textit{data mining} and \textit{database} are related fields, but they cannot understand how exactly they are related, e.g, \textit{why is it a facet?} and \textit{what is the facet?}

Although techniques like knowledge graph reasoning \cite{lao2011random,xiong2017deeppath,chen2018variational} or open relation extraction \cite{etzioni2008open} can represent more complex relationships to some extent, they do not fundamentally solve the limitations as discussed in \citet{huang2022open}. For instance, neither a multi-hop reasoning path in KGs nor a triple extracted by open relation extraction, e.g., (data mining methods, \textit{to be integrate within}, the framework of traditional database systems), is easy to interpret.

Based on the above analysis, we propose a new form of modeling relationships between entities: \deer{}\deeremoji{} (\textbf{D}escriptive Knowledge Graph for \textbf{E}xplaining \textbf{E}ntity \textbf{R}elationships). We define \deer{} as a graph, where nodes are entities and edges are descriptive statements of entity relationships (refer to Figure \ref{fig:graph} for an example). \deer{} is \textbf{\textit{open}} since it
does not require a pre-specified set of relation types. In principle, all entity relationships, either explicit or implicit, can be represented by \deer{}, as long as they can be connected in a sentence -- which is not possible for KGs. It is \textbf{\textit{informative}} since the relationships between entities are represented by informative free-text relation descriptions, instead of simple short phrases like ``facet of''. 

\deer{} has great potential to help users understand entity relationships more easily and intuitively by providing relation descriptions for any two related entities and facilitate downstream tasks on entities and entity relationships such as entity profiling \cite{noraset2017definition,cheng2020ent,huang2021understanding}, relation extraction \cite{bach2007review}, and knowledge graph completion \cite{lin2015learning}.
For example, in Figure \ref{fig:graph}, we can understand the semantic meaning of the terms by connecting them with familiar ones.
In e-commerce, the system (e.g., Amazon online shopping website) may recommend \textit{tripods} to a photography novice who is browsing \textit{cameras}.
An explanation in \deer{}, e.g., ``\textit{tripods} are used for both motion and still photography to prevent \textit{camera} movement and provide stability'', could not only help users make a better purchase decision but also justify the recommendation. In KG construction and completion, the relation descriptions can serve as knowledge to improve the performance or as explanations to justify the relations in KGs.

The key to building \deer{} is to acquire high-quality relation descriptions.
However, writing or collecting relation descriptions manually requires enormous human efforts and expertise (in our human evaluation in Section \ref{sec:exp_construction}, it takes $\sim$3 minutes to evaluate whether a sentence is a good relation description). 
Considering this, we propose a novel two-step approach to construct \deer{} with Wikipedia, where no manual annotation is required.  
Specifically, we first \textbf{\textit{extract}} relation descriptions from corpus in a self-supervised manner, where a scoring function is introduced to measure the \textit{explicitness}, i.e., how explicit is the relationship represented by the sentence, and \textit{significance}, i.e., how significant is the relationship represented, with the analysis of dependency patterns.
Second, based on the extracted graph,
a transformer-based relation description synthesizing model is introduced to \textbf{\textit{generate}} relation descriptions for interesting entity pairs whose relation descriptions are not extracted in the first step. 
This allows \deer{} to handle a large number of entity pairs, including those that do not co-occur in the corpus.

Both quantitative and qualitative experiments demonstrate the effectiveness of our proposed methods.
We also conduct case study and error analysis and suggest several promising directions for future work -- \deer{} not only serves as a valuable application in itself to help understand entity relationships,
but also has the potential to serve as a knowledge source to facilitate various tasks on entities and entity relationships.

\section{Related Work}

There are several previous attempts on acquiring entity relation descriptions. For instance, \citet{voskarides2015learning} study a learning to rank problem of ranking relation descriptions by training a Random Forest classifier with manually annotated data. Subsequently, \citet{huang2017learning} build a pairwise ranking model based on convolutional neural networks by leveraging query-title pairs derived from clickthrough data of a Web search engine, and \citet{voskarides2017generating} attempt to generate descriptions for relationship instances in KGs by filling created sentence templates with appropriate entities. 
However, all these methods are not ``open''. First, they rely and demand heavily on features of entities and relations. Second, these models only deal with entities with several pre-specified relation types, e.g., 9 in \citet{voskarides2015learning} and 10 in \citet{voskarides2017generating}, and only explicit relation types, e.g., \textit{isMemberOfMusicGroup}, are covered.
Notably, \citet{handler2018relational} propose to extract relation statements, i.e., natural language expressions that begin with one entity and end with the other entity, from a corpus to describe entity relationships. However, the ``\textit{acceptability}'' used in their work cannot ensure a good relation description.
Moreover, these works do not systematically analyze and define what constitutes a good relational description.

The work most relevant to ours is \textit{Open Relation Modeling} \cite{huang2022open}, which aims to generate relation descriptions for entity pairs. To achieve this, the authors propose to fine-tune BART \cite{lewis2020bart} to reproduce definitions of entities. 
Compared to their problem, i.e., text generation, the focus of this paper is on graph construction.
Besides, their relation descriptions are limited to definitional sentences, which assumes that one entity appears in the other's definition; however, the assumption is not true for many related entities. In addition, their methodology does not incorporate sufficient knowledge about entities and relations for generation. 

There are also some other works that can be related. For example, \citet{lin2020commongen,liu2021kg} study \textit{CommonGen}, which aims to generate coherent sentences containing the given common concepts.
\citet{dognin2020dualtkb,agarwal2021knowledge} study the data-to-text generation \cite{kukich1983design}, which aims to convert facts in KGs into natural language. \citet{gunaratna2021entity} propose to construct an entity context graph with contexts as random paragraphs containing the target entities to help entity embedding. 
None of them meets the requirements for high-quality relation descriptions.

\section{Descriptive Knowledge Graph for Explaining Entity Relationships}
\label{sec:DKG}

\deer{}\deeremoji{} is a graph representing entity relationships with sentence descriptions. Formally, we define \deer{} as a directed graph $\mathcal{G} = \{ \mathcal{E}, \mathcal{R} \}$, where $\mathcal{E}$ is the set of entities and $\mathcal{R}$ is the set of relation description facts. A relation description fact is a triple $(x,s,y)$, where $x, y \in \mathcal{E}$ are the \textit{subject} and \textit{object} of $s$, respectively. $s$ is a sentence describing the relationship between $x$ and $y$ (Figure \ref{fig:graph}).

To build \deer{}, the first step is to collect entities and identify related entity pairs, which can be simply achieved by utilizing existing resources, e.g., Wikipedia, and entity relevance analysis, e.g., cosine similarity of entity embeddings in Wikipedia2vec \cite{yamada2020wikipedia2vec}.
And then, we need to acquire high-quality relation descriptions for entity pairs.
Taking entity pair (\textit{machine learning}, \textit{algorithm}) as an example, a relation description of them can be
$s_1$ in Table \ref{table:example}.
From the perspective of human understanding,
we identify three requirements for a good relation description:
\begin{itemize}[noitemsep,nolistsep,leftmargin=*]
    \item \textbf{Explicitness}: The relationship of the target entities is described explicitly. E.g., in $s_1$, ``\textit{machine learning explores the study and construction of algorithms}'' describes the relationship explicitly; while in $s_2$, the relationship between \textit{machine learning} and \textit{algorithm} is expressed implicitly so that the relationship is difficult to reason. 
    \item \textbf{Significance}: The relationship of the target entities is the point of the sentence. In $s_1$, all the tokens in the sentence are associated with the relationship between \textit{machine learning} and \textit{algorithm}; while in $s_3$, although the description is explicit, ``\textit{which
    ... far}'' mainly characterizes \textit{algorithm}, but not the target entity relationship.
    \item \textbf{Correctness}: The relationship between target entities is described correctly.
\end{itemize}

\begin{table}[tp!]
\scriptsize
\setlength\tabcolsep{1.5pt}
\begin{tabular}{c|p{0.92\linewidth}}
\toprule
\# & \multicolumn{1}{c}{\textbf{Sentence}} \\
\midrule[1pt]
$s_1$ & \emph{Machine learning} explores the study and construction of \emph{algorithms} that can learn from and make predictions on data.\\
\hline
$s_2$ & \emph{Machine learning} is employed in a range of computing tasks where designing and programming explicit, rule-based \emph{algorithms} is infeasible. \\
\hline
$s_3$ & \emph{Machine learning} includes \emph{algorithms} that are adaptive or have adaptive variants, which usually means that the algorithm parameters are automatically adjusted according to statistics about the optimisation thus far. \\
\bottomrule
\end{tabular}
\vspace{-1mm}
\caption{Example sentences containing both \textit{machine learning} and \textit{algorithm}.}
\label{table:example}
\vspace{-5mm}
\end{table}

There are other requirements to ensure a good relation description, e.g., the sentence is coherent, grammatical, of reasonable length. 
Compared to the above ones, these requirements are general requirements for any sentence, but not specific to our problem;
therefore, we put less emphasis on them.

To acquire relation descriptions that satisfy the above requirements, we propose a novel two-step approach: 
first extracting relation descriptions from a corpus with the analysis of dependency patterns (Section \ref{sec:extraction}), 
and then generating relation descriptions for interesting entity pairs whose relation descriptions are not extracted in the previous step (Section \ref{sec:generation}).

\section{Relation Description Extraction}
\label{sec:extraction}

In this section, we introduce our approach for extracting entity relation descriptions from Wikipedia according to the requirements discussed in Section~\ref{sec:DKG}.

\subsection{Preprocessing and Filtering}
\label{sec:preprocessing}

The goal of preprocessing and filtering is to collect entities and map entity pairs to candidate relation descriptions.
To ensure correctness, we use Wikipedia as the source corpus, which is a high-quality corpus covering a wide range of domains.
Because this process mainly relies on heuristic rules and existing tools, to save space, we refer the readers to Appendix \ref{app:preprocessing} for the details.

\begin{figure*}[tp!]
\centerline{\includegraphics[width=\linewidth]{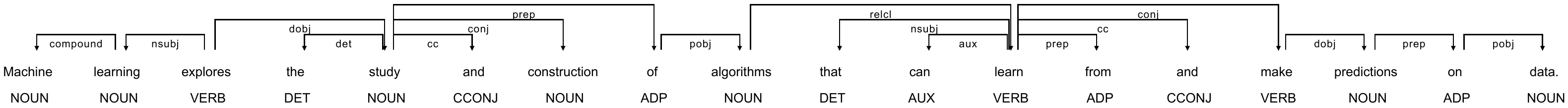}}
\vspace{-1mm}
\caption{Dependency tree of $s_1$.}
\vspace{-4mm}
\label{fig:dep_s1}
\end{figure*}

\subsection{Scoring}

In this section, we design a scoring function to measure the quality of relation descriptions.
Since we use Wikipedia as the source corpus, the \textit{correctness} of the extracted sentences can be largely guaranteed;
thus, we focus on measuring \textit{explicitness} and \textit{significance} of candidate relation descriptions.

\subsubsection{Shortest Dependency Path as Relation}

Inspired by \citet{wu2010open}, we use the shortest dependency path to represent the relation pattern between the target entities in a sentence. 
For instance, Figure \ref{fig:dep_s1} shows the dependency tree of $s_1$ processed by spaCy\footnote{\url{https://spacy.io}}. 
The shortest path between \textit{machine learning} and \textit{algorithm} is:
``learning $\overleftarrow{nsubj}$ explores $\overrightarrow{dobj}$ study $\overrightarrow{prep\vphantom{b}}$ of $\overrightarrow{pobj}$ algorithms''.
Following their notation, we call such a path a \textit{corePath}. 
To represent the relation pattern, we collect dependencies in the path and append ``i\_'' to the dependencies with an inversed direction. E.g., the relation pattern for the above path is [$i\_nsubj, dobj, prep, pobj$].
We remove dependencies that do not affect human understanding. Specifically, we drop the \textit{conj} and \textit{appos} dependencies and replace two consecutive \textit{prep} with one.

Besides \textit{corePath}, we also collect the shortest paths between the \textit{corePath} and the tokens outside the \textit{corePath} to represent the relationships between entity relationships and tokens. 
For instance, in Figure \ref{fig:dep_s1}, \textit{construction} is a token outside the \textit{corePath} between \textit{machine learning} and \textit{algorithm}. The shortest path between it and the \textit{corePath} is:
``study $\overrightarrow{conj}$ construction''. We call this kind of path as \textit{subPath}. Similar to \textit{corePath}, we generate the relation pattern from \textit{subPath} and drop the \textit{conj}, \textit{appos} and \textit{compound} dependencies.

\subsubsection{Explicitness}
\label{sec:explicitness}

Given two entities and a candidate relation description $s$, we measure the explicitness by calculating the normalized logarithmic frequency of the relation pattern of the \textit{corePath}:
\begin{equation}
\label{eq:E}
ExpScore(s) = \frac{\log(f_p+1)}{\log(f_{max}+1)},
\end{equation}
where $f_{max}$ is the frequency of the most frequent \textit{corePath} relation pattern and $f_p$ is the frequency of the relation pattern in the present \textit{corePath}. The intuition here is that humans tend to use explicit structure to explain relations. Thus, we assume that a relation description is more explicit if its relation pattern is more frequent. Intuitively, if a relation pattern is unpopular, it is likely that this pattern is either too complicated or contains some rarely used dependencies. Both of these cases may increase the difficulty in reasoning.

Similar to \citet{wu2010open},
we only consider patterns that start with \textit{nsubj} or \textit{nsubjpass}, indicating that one of the target entities is the subject of the sentence. This restriction helps increase the explicitness of the selected relation description sentences because if one entity is the subject, the sentence is likely to contain a ``argument-predicate-argument'' structure connecting the target entities.

\subsubsection{Significance}

We measure the significance as the proportion of information that is relevant to the entity relationship in a sentence. 
To measure the relevance of each token in the sentence to the entity relationship, we divide tokens into three categories: 1) \textit{core token} if the token is in the \textit{corePath}; 2) \textit{modifying token} if the token is in a \textit{subPath} that is connected to the \textit{corePath} through a modifying dependency; and 3) \textit{irrelevant token} for the rest tokens.
The intuition here is that a sub-dependency tree connected to the \textit{corePath} with a modifying dependency is supposed to modify the relationship. We predefined a set of modifying dependencies in Table \ref{table:modifying_dependencies}.

We calculate a score for each token in the sentence based on its category and dependency analysis. Then, the significance score is the average of all the token's scores. 
Formally, for a candidate relation description $s$, the significance score is
\begin{equation}
\label{eq:S}
SigScore(s) = \frac{\sum_{t \in s}{w(t)}}{|s|},
\end{equation}
where
\begin{equation}
w(t) = 
\begin{cases}
1 & \text{if $t \in ct$} \\
\frac{\log(f'_{p_t}+1)}{\log(f'_{max}+1)}  & \text{if $t \in mt$} \\
0 & \text{otherwise}
\end{cases},
\end{equation}
where $ct$ is the set of \textit{core tokens} and $mt$ is the set of \textit{modifying tokens}.
$f'_{p_t}$ is the frequency of the \textit{subPath} relation pattern from the \textit{corePath} to the present token $t$ and $f'_{max}$ is the frequency of the most frequent \textit{subPath} relation pattern. The intuition is: with higher relation pattern frequency, the modifying token is more explicitly related to the entity relationship, and thus, should have a higher score. This also comes with another useful characteristic: the score will decrease token by token as we move along the \textit{subPath} because the frequency of a \textit{subPath} relation pattern cannot be greater than the frequency of its parent. With this characteristic, we can penalize the long modifying \textit{subPath} as it will distract the focus from the entity relationship and is less explicitly related to the relationship.

\subsubsection{Relation Descriptive Score}

To calculate the explicitness and significance, we need to build a database of relation patterns for both \textit{corePath} and \textit{subPath}. We construct both databases with the candidate relation descriptions and corresponding entity pairs collected from Section \ref{sec:preprocessing} with spaCy. 
We also require the two target entities in the sentence are related to a certain threshold. 
Intuitively, if two entities are more related, the sentences containing them are more likely to be relation descriptions; therefore, the extracted \textit{corePath} relation patterns are more likely to indicate entity relationships.
We measure the relevance of two entities by calculating the cosine similarity of the entity embeddings in Wikipedia2Vec. We filter out entity pairs (and the associated sentences) with a relevance score $< 0.5$.
This leads to a collection of 7,186,996 \textit{corePaths} and 83,265,285 \textit{subPaths}.

With the databases of relation patterns, we can calculate the explicitness and significance scores for a candidate relation description. The final score, named \textbf{\textit{Relation Descriptive Score (RDScore)}}, is computed as the harmonic mean:
\begin{equation}
\label{eq:RDScore}
RDScore(s) = 2 \cdot \frac{ExpScore(s) \cdot SigScore(s)}{ExpScore(s) + SigScore(s)}.
\end{equation}

For each entity pair,
we calculate \textit{RDScore} for all the candidate relation descriptions and select the candidate with the highest score as the final relation description. 
To build an initial \deer{}, we keep edges with an entity relevance score $\geq 0.5$\footnote{Since there is no boundary that delineates whether two entities are related, we consider the relevance threshold as a hyperparameter.}
and with a relation description whose \textit{RDScore} $\geq 0.75$\footnote{This threshold is also a hyperparameter to balance the density of the graph and the quality of relation descriptions.}. 
We refer to this graph as \textbf{Wiki-\deer{}$_0$}.

\section{Relation Description Generation}
\label{sec:generation}

\begin{figure*}[tp!]
\centerline{\includegraphics[width=0.95\linewidth]{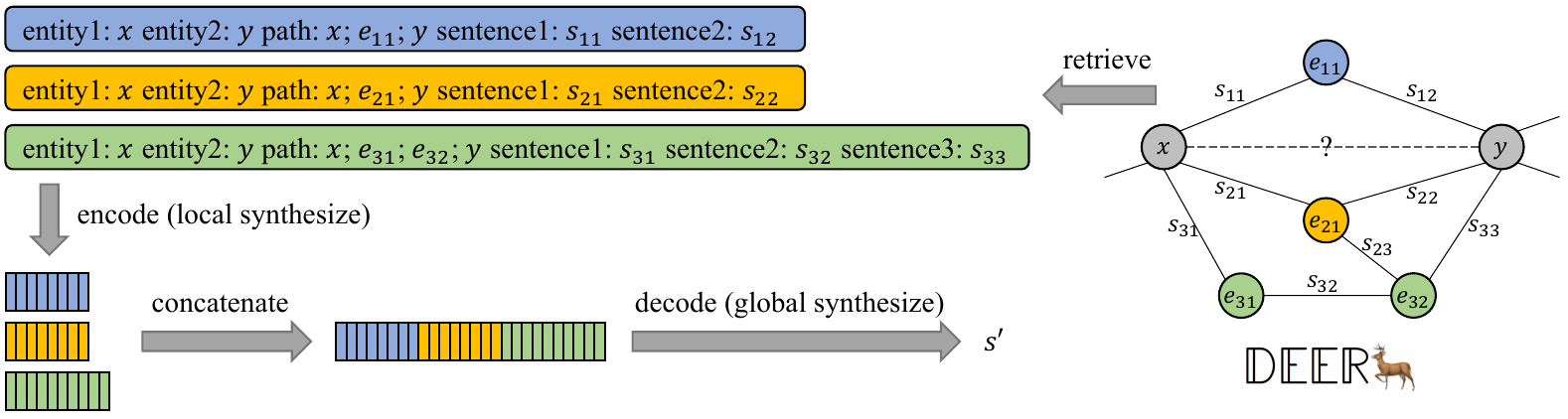}}
\vspace{-1mm}
\caption{The framework of \textit{RelationSyn}. Given entity pair $(x, y)$ whose relation description is not present in the initial \deer{}, we first retrieve several reasoning paths from the graph. And then, we encode (local synthesize) each reasoning path into a latent vector and concatenate all the latent vectors. Finally, we decode (global synthesize) the vector to produce relation description $s'$ for $(x, y)$.}
\vspace{-4mm}
\label{fig:RelationSyn}
\end{figure*}

In the previous section, we extract relation descriptions for entity pairs with the analysis of dependency patterns and build an initial \deer{} with Wikipedia automatically. 
However, for some related entity pairs, there may not exist a sentence that contains both entities; and although such a sentence exists, it may not be extracted by the system.
To solve this problem, in this section, we introduce \textit{Relation Description Generation} -- generating relation descriptions for interesting entity pairs.

\label{sec:relationsyn}

We form relation description generation as a conditional text generation task: given two entities, generating a sentence describing the relationship between them with the initial \deer{}. Formally, we apply the knowledge-enhanced sequence-to-sequence formulation \cite{yu2020survey}: given an entity pair $(x,y)$ and an initial \deer{} $\mathcal{G}_0$, the probability of the output relation description $s$ is computed auto-regressively:
\begin{equation}
P(s|x, y, \mathcal{G}_0) = \prod_{i=1}^{m} P(s_i|s_{0:i-1}, x, y, \mathcal{G}_0),
\end{equation}
where $m$ is the length of $s$, $s_i$ is the $i$th token of $s$, and $s_0$ is a special start token.

To incorporate $\mathcal{G}_0$ for generation, we propose \textit{\textbf{Relation Description Synthesizing (RelationSyn)}}. RelationSyn consists of two processes: first retrieving relevant relation descriptions (reasoning paths) from the graph and then synthesizing them into a final relation description (Figure \ref{fig:RelationSyn}). 

\subsection{Retrieval}
\label{sec:retrieval}

To generate a relation description, the model needs knowledge about the target entities and their relationship. To provide knowledge, 
we retrieve reasoning paths of the target entities from the graph. 

In \deer{}, we define a reasoning path $q$ as a path connecting the target entities, which is called $k$-hop if it is connected by $k$ edges.
For instance, in Figure \ref{fig:RelationSyn}, there are two $2$-hop reasoning paths between $x$ and $y$: $(x,s_{11},e_{11},s_{12},y)$ and $(x,s_{21},e_{21},s_{22},y)$, and two $3$-hop reasoning paths: $(x,s_{21},e_{21},s_{23},e_{32},s_{33},y)$ and $(x,s_{31},e_{31},s_{32},e_{32},s_{33},y)$ in the graph\footnote{In order to collect more reasoning paths as knowledge for generation, we ignore the directions of edges.}. 
To measure the quality of reasoning paths, we define \textit{PathScore} as the harmonic mean of \textit{RDScore} of relation descriptions in the path:
\begin{equation}
PathScore(q) = \frac{|\mathcal{S}_q|}{\sum_{s \in \mathcal{S}_q}{\frac{1}{RDScore(s)}}},
\end{equation}
where $\mathcal{S}_q$ is the set of relation descriptions in $q$, and $|\mathcal{S}_q|=k$.

Reasoning paths are helpful for relation description generation. For instance, 
from reasoning path $(\text{deep learning},$ $s_1',$ $\text{machine learning}, s_2',$ $\text{artificial intelligence})$ (refer to Figure \ref{fig:graph} for $s_1'$ and $s_2'$), we can infer the relationship between \textit{deep learning} and \textit{AI}: \textit{deep learning} is the dominant approach for \textit{ML}, while \textit{ML} grew out of the quest for \textit{AI}; therefore, \textit{deep learning} is an important technology for the development of \textit{artificial intelligence}.

However, not all reasoning paths are equally useful. 
Longer reasoning paths are usually more difficult to reason, while paths with higher \textit{PathScore} usually contain more explicit and significant relation descriptions.
Therefore, when retrieving reasoning paths for an entity pair, we first sort the paths by their length (shorter first) and then by their \textit{PathScore} (higher first).

\subsection{Synthesizing}

According to Section \ref{sec:retrieval}, we may retrieve multiple reasoning paths for an entity pair whose relation description is missed in the initial \deer{}.
In this section, we focus on synthesizing relation descriptions in the retrieved reasoning paths into a final relation description of the target entities based on T5 \cite{raffel2020exploring} and Fusion-in-Decoder \cite{izacard2021leveraging}. 

We first convert each reasoning path to a sequence using the following encoding scheme: e.g., $(x,s_{31},e_{31},s_{32},e_{32},s_{33},y)$ $\to$ ``entity1: $x$ entity2: $y$ path: $x$; $e_{31}$; $e_{32}$; $y$ sentence1: $s_{31}$ sentence2: $s_{32}$ sentence3: $s_{33}$''. And then, we encode the sequence with the encoder of T5. In this way, the relation descriptions in each reasoning path are synthesized into a latent vector, named ``\textbf{local synthesizing}''.

After local synthesizing, we concatenate the latent vectors of all the retrieved reasoning paths to form a global latent vector. The decoder of T5 performs attention over the global latent vector and produces the final relation description. We name this process as ``\textbf{global synthesizing}''.

Combining retrieval and synthesizing, given two entities, we first retrieve $m$ reasoning paths connecting the target entities according to their length and \textit{PathScore}, and then synthesize them to produce the target relation description. We refer to this model as \textbf{RelationSyn-$m$}.

\section{Evaluation}

In this section, we verify the proposed methods for building \deer{} by conducting experiments on relation description extraction and generation.

\subsection{Relation Description Extraction}
\label{sec:exp_construction}

We first present the statistics of the initial \deer{} built with Wikipedia in Table \ref{table:Wiki-DKG}.

To evaluate the quality of relation descriptions in the graph, we randomly sample 100 entity pairs from the graph\footnote{More specifically, for better comparison with generation later, we sample 100 entity pairs from the test set in Table \ref{table:data_completion}.} and ask three human annotators (graduate students doing research on computational linguistics) to assign a graded value (1-5) for each relation description according to Table \ref{table:rating_scale}.

Since previous works on relation description extraction are supervised and only limited to several explicit relation types, e.g., 9 in \citet{voskarides2015learning}, it is impractical and meaningless to compare with them. For instance, the relationship of (\textit{Arthur 
Samuel}, \textit{Machine Learning}) is not available or even not considered by the previous methods. Therefore, we verify the effectiveness of our model by comparing different variants of the model:
\begin{itemize}[noitemsep,nolistsep,leftmargin=*]
    \item \textbf{Random}: A sentence containing the target entities is randomly selected as the relation description.
    \item \textbf{ExpScore}: The sentence with the highest \textit{explicitness} is selected according to Eq. \eqref{eq:E}.
    \item \textbf{SigScore}: The sentence with the highest \textit{significance} is selected according to Eq. \eqref{eq:S}.
    \item \textbf{RDScore}: The sentence with the highest \textit{RDScore} is selected according to Eq. \eqref{eq:RDScore}.
\end{itemize}

\begin{table}[tp!]
    \small
    \begin{center}
    \begin{tabular}{r|r|c}
        \toprule
        \textbf{\# nodes} & \textbf{\# edges} & \textbf{average sentence length} \\
        \midrule
        1,378,471 & 2,890,718 & 19.9\\
        \bottomrule
    \end{tabular}
    \end{center}
    \vspace{-1mm}
    \caption{The statistics of Wiki-\deer{}$_0$.}
    \vspace{-3mm}
    \label{table:Wiki-DKG}
\end{table}

\begin{table}[tp!]
\small
\begin{center}
\begin{tabular}{l|c}
\toprule
 & \textbf{Rating (1-5)} \\
\midrule 
Random & 2.75 \\
\hline
ExpScore & 3.77 \\
SigScore & 3.84 \\
\hline
RDScore & \textbf{4.18} \\
\bottomrule
\end{tabular}
\vspace{-1mm}
\caption{Qualitative results of extraction.}
\vspace{-2mm}
\label{table:human_evaluation_extracion}
\end{center}
\end{table}

Table \ref{table:human_evaluation_extracion} shows the human evaluation results for relation description extraction, with an average pairwise
Cohen's $\kappa$ of 0.66 (good agreement).
From the results, we observe that both our explicitness and significance measurements are important to ensure a good relation description. 
In addition, \textit{RDScore} achieves an average rating of 4.18, which means that most of the selected sentences are high-quality relation descriptions, further indicating that the quality of Wiki-\deer{}$_0$ is high.

\subsection{Relation Description Generation}
\label{sec:exp_completion}

\subsubsection{Experimental Setup}

{\flushleft \textbf{Data construction}.} 
We build a dataset for relation description generation as follows: 
for an entity pair with a relation description in Wiki-\deer{}$_0$, we hide the relation description and consider it as the target for generation. The goal is to recover/generate the target relation description with the rest of the graph\footnote{To increase the difficulty of the task, we assume these two entities do not co-occur in the corpus, i.e., we do not utilize any sentence containing both the target entities for generation.}.
For instance, in Figure \ref{fig:RelationSyn}, we hide the edge (relation description $s$) between $x$ and $y$ and use the remaining reasoning paths to recover $s$.
We train and test on entity pairs with $\geq5$ reasoning paths connecting them. The statistics of the data are reported in Table \ref{table:data_completion}.

\begin{table}[tp!]
    \small
    \begin{center}
    \begin{tabular}{c|r|r|r}
        \toprule
        & \textbf{train} & \textbf{valid} & \textbf{test} \\
        \midrule
        size & 847,792 & 17,662 & 17,663 \\
        \bottomrule
    \end{tabular}
    \end{center}
    \vspace{-1mm}
    \caption{The statistics of data for generation.}
    \vspace{-3mm}
    \label{table:data_completion}
\end{table}

\begin{table*}[tp!]
\begin{center}
\small
\begin{tabular}{l|c|c|c|c}
\toprule
 & \textbf{BLEU} & \textbf{ROUGE} & \textbf{METEOR} & \textbf{BERTScore} \\
\midrule 
RealtionBART-Vanilla \cite{huang2022open} & 19.61 & 41.52 & 20.48 & 82.99 \\
RealtionBART-MP + PS \cite{huang2022open} & 21.64 & 42.62 & 21.40 & 83.29 \\
\hline
RelationSyn-$0$ & 20.83 & 41.46 & 20.66 & 82.84 \\
\hline
\textbf{RelationSyn-$1$} & 22.43 & 42.74 & 21.65 & 83.41  \\
\textbf{RelationSyn-$3$} & 23.26 & 43.33 & 22.12 & 83.63  \\
\textbf{RelationSyn-$5$} & \textbf{23.88} & \textbf{43.56} & \textbf{22.40} & \textbf{83.70}  \\
\bottomrule
\end{tabular}
\vspace{-1mm}
\caption{Quantitative results of relation description generation.}
\vspace{-3mm}
\label{table:quan_completion}
\end{center}
\end{table*}

{\flushleft \textbf{Models}.} The task of relation description generation is relevant to \textit{Open Relation Modeling} \cite{huang2022open} -- a recent work aimed at generating sentences capturing general relations between entities conditioned on entity pairs. To the best of our knowledge, no other existing work can generate relation descriptions for any two related entities (since open relation modeling has only just been introduced). Therefore, we mainly compare the models proposed in \citet{huang2022open} 
with several variants of our model:
\begin{itemize}[noitemsep,nolistsep,leftmargin=*]
    \item \textbf{RelationBART (Vanilla)}: The vanilla model proposed in \citet{huang2022open} for generating entity relation descriptions, where BART \cite{lewis2020bart} is fine-tuned on a training data whose inputs are entity pairs and outputs are corresponding relation descriptions.
    \item \textbf{RelationBART-MP + PS}: The best model proposed in \citet{huang2022open}, which incorporates Wikidata by selecting the most interpretable and informative reasoning path in the KG automatically for helping generate relation descriptions.
    \item \textbf{RelationSyn-$0$}: A reduced variant of our model, where the encoding scheme of the input is only ``entity1: $x$ entity2: $y$'', i.e., no reasoning path and relation description is fed to the encoder. 
    \item \textbf{RelationSyn-$m$}: The proposed relation description synthesizing model (Section \ref{sec:relationsyn}), where $m$ is the maximum number of retrieved reasoning paths for an entity pair.
\end{itemize}

{\flushleft \textbf{Metrics}.} We perform both quantitative and qualitative evaluation. Following \citet{huang2022open}, we apply several automatic metrics, including BLEU \cite{papineni2002bleu}, ROUGE-L \cite{lin2004rouge}, METEOR \cite{banerjee2005meteor}, and BERTScore \cite{zhang2019bertscore}.
Among them, BLEU, ROUGE, and METEOR focus on measuring surface similarities between the generated relation descriptions and the target relation descriptions, and BERTScore is based on the similarities of contextual token embeddings.
We also ask three human annotators to evaluate the output relation descriptions with the same rating scale in Table \ref{table:rating_scale}.

{\flushleft \textbf{Implementation details}.}
We train and evaluate all the baselines and variants on the same train/valid/test split. For \textit{RelationBART (Vanilla)} and \textit{RelationBART-MP + PS}, we apply the official implementation\footnote{\url{https://github.com/jeffhj/open-relation-modeling}} and adopt the default hyperparameters. The training converges in 50 epochs. For our models, we modify the implementation of Fusion-in-Decoder\footnote{\url{https://github.com/facebookresearch/FiD}} and initialize the model with the T5-base configuration. All the baseline models for \textit{RelationSyn} are trained under the same batch size of 8 with a learning rate of $0.0001$ and evaluated on the 
validation set every 5000 steps. The training is considered converged and terminated with no better performance on the validation set in 20 evaluations. The training of all models converges in 20 epochs. The training time is about one week on a single NVIDIA A40 GPU. For evaluation, the signature of BERTScore is: roberta-large-mnli L19 no-idf version=0.3.11(hug trans=4.15.0).

\begin{table}[tp!]
\small
\begin{center}
\begin{tabular}{l|c}
\toprule
 & \textbf{Rating (1-5)} \\
\midrule 
\textit{Random} & 2.75 \\
\textit{RDScore (Oracle)} & 4.18 \\
\hline
RealtionBART-MP + PS & 3.12 \\
RelationSyn-$0$ & 3.08 \\
\textbf{RelationSyn-$1$} & 3.34 \\
\textbf{RelationSyn-$5$} & \textbf{3.47} \\
\bottomrule
\end{tabular}
\vspace{-1mm}
\caption{Qualitative results of generation.}
\vspace{-4mm}
\label{table:qual_completion}
\end{center}
\end{table}

\subsubsection{Quantitative Evaluation}

Table \ref{table:quan_completion} reports the results of relation description generation with the automatic metrics. We observe that our best model \textit{RelationSyn-$5$} outperforms the state-of-the-art model for open relation modeling significantly. We also observe that \textit{RelationSyn-$1$} performs better than \textit{RelationSyn-$0$}, which means that reasoning paths in \deer{} are helpful for relation description generation. In addition, as the number of reasoning paths, i.e., $m$, increases, the performance of RelationSyn-$m$ improves. This demonstrates that the proposed model can synthesize multiple relation descriptions in different reasoning paths into a final relation description.

\subsubsection{Qualitative Evaluation}

We also conduct qualitative experiments to measure the quality of generated relation descriptions. For a better comparison with extraction, we sample the same 100 entity pairs from the test set as in Section \ref{sec:exp_construction}. 
From the results in Table \ref{table:qual_completion}, we observe that the quality of generated relation descriptions is higher than that of random sentences containing the target entities.
The best model, \textit{RelationSyn-$5$}, achieves a rating of 3.47, which means the model can generate reasonable relation descriptions. However, the performance is still much worse than \textit{Oracle}, i.e., relation descriptions extracted by our best extraction model (\textit{RDScore}). 
This indicates that generating high-quality relation descriptions is still a challenging task.

\subsection{Case Study and Error Analysis}
\label{sec:analysis}

In Table \ref{table:case} of Appendix \ref{app:examples}, we show some sample outputs in the test set of relation description generation of three extraction models: \textit{ExpScore}, \textit{SigScore}, \textit{RDScore}, and three generation models: \textit{RelationSyn-$0$}, \textit{RelationSyn-$1$}, \textit{RelationSyn-$5$}.

For extraction, we observe that if we only consider the explicitness of the sentence, the selected sentence may contain a lot of stuff that is irrelevant to the entity relationship, e.g., (\textit{Mucus}, \textit{Stomach}). And if we only consider the significance, the relationship between entities may be described implicitly; thus the relationship is difficult to reason out, e.g., (\textit{Surfers Paradise}, \textit{Queensland}) and (\textit{Knowledge}, \textit{Epistemology}). And the combination of them, i.e., \textit{RDScore}, yields better relation descriptions.

For generation, we notice that \textit{RelationSyn-$0$} suffers severely from \emph{hallucinations}, i.e., generating irrelevant or contradicted facts. E.g., the relation descriptions generated for (\textit{Dayan Khan}, \textit{Oirats}) is incorrect. By incorporating relation descriptions in the reasoning paths as knowledge, hallucination is alleviated to some extent, leading to better performance of \textit{RelationSyn-$1$} and \textit{RelationSyn-$5$}.

From the human evaluation results, we also find that the correctness of relation descriptions extracted by \textit{RDScore} is largely guaranteed. However, sometimes, the extracted sentences are still a bit implicit or not significant. In contrast to this, the relation descriptions generated by \textit{RelationSyn} are usually explicit and significant (the average \textit{RDScore} of the relation descriptions generated by \textit{RelationSyn-$5$} is 0.886, compared to 0.853 of \textit{Oracle}), but contain major or minor errors. We think this is because most of the relation descriptions extracted by \textit{RDScore} are explicit and significant, and the generation model can mimic the dominant style of relation descriptions in the training set. 
However, it is still challenging to generate fully correct relation descriptions by synthesizing existing relation descriptions.

We also attempted to find the eight entity pairs in Table \ref{table:case} in Wikidata.
Among them, only (\textit{Surfers Paradise}, \textit{Queensland}) is present in Wikidata.
This further confirms that \deer{} can model a wider range of entity relationships.

\section{Conclusion and Discussion}

In this paper, we propose \deer{}\deeremoji{} -- an open and informative form of modeling relationships between entities. To avoid tremendous human efforts, we design a novel self-supervised learning approach to extract relation descriptions from Wikipedia.
To provide relation descriptions for related entity pairs whose relation descriptions are not extracted in the previous step, we study relation description generation by synthesizing relation descriptions in the retrieved reasoning paths.
We believe that \deer{} can not only serve as a direct application to help understand entity relationships but also be utilized as a knowledge source to facilitate related tasks such as relation extraction \cite{bach2007review} and knowledge graph completion \cite{lin2015learning}.

\section*{Limitations}

In this paper, we focus on designing methods to construct \deer{} and evaluating \deer{} on serving as a system for entity relationship understanding, which has direct applications in, e.g., encyclopedias and concept maps.
Due to limited space, we do not fully investigate its use as a knowledge source to facilitate other tasks, e.g., relation extraction and knowledge graph completion, which we leave as future work for the whole research community.

\section*{Acknowledgements}

We thank the reviewers for their constructive feedback.
This material is based upon work supported by the National Science Foundation IIS 16-19302 and IIS 16-33755, Zhejiang University ZJU Research 083650, IBM-Illinois Center for Cognitive Computing Systems Research (C3SR) -- a research collaboration as part of the IBM Cognitive Horizon Network, grants from eBay and Microsoft Azure, UIUC OVCR CCIL Planning Grant 434S34, UIUC CSBS Small Grant 434C8U, and UIUC New Frontiers Initiative. Any opinions, findings, and conclusions or recommendations expressed in this publication are those of the author(s) and do not necessarily reflect the views of the funding agencies.

\bibliography{anthology,custom}
\bibliographystyle{acl_natbib}

\clearpage

\appendix

\begin{table}[tp!]
\scriptsize
\begin{center}
\begin{tabular}{l|c}
\toprule
\textbf{Dependency label} & \textbf{Description} \\
\midrule 
acl & clausal modifier of noun (adjectival clause) \\
\hline
advcl & adverbial clause modifier \\
\hline
advmod & adverbial modifier \\
\hline
amod & adjectival modifier \\
\hline
det & determiner \\
\hline
mark & marker \\
\hline
meta & meta modifier \\
\hline
neg & negation modifier \\
\hline
nn & noun compound modifier \\
\hline
nmod & modifier of nominal \\
\hline
npmod & noun phrase as adverbial modifier \\
\hline
nummod & numeric modifier \\
\hline
poss & possession modifier \\
\hline
prep & prepositional modifier \\
\hline
quantmod & modifier of quantifier \\
\hline
relcl & relative clause modifier \\
\hline
appos & appositional modifier \\
\hline
aux & auxiliary \\
\hline
auxpass & auxiliary (passive) \\
\hline
compound & compound \\
\hline
cop & copula \\
\hline
ccomp & clausal complement \\
\hline
xcomp & open clausal complement \\
\hline
expl & expletive \\
\hline
punct & punctuation \\
\hline
nsubj & nominal subject \\
\hline
csubj & clausal subject \\
\hline
csubjpass & clausal subject (passive) \\
\hline
dobj & direct object \\
\hline
iobj & indirect object \\
\hline
obj & object \\
\hline
pobj & object of preposition \\
\bottomrule
\end{tabular}
\caption{Manually collected modifying dependencies in spaCy.}
\vspace{-4mm}
\label{table:modifying_dependencies}
\end{center}
\end{table}

\section{Preprocessing and Filtering}
\label{app:preprocessing}

We introduce our preprocessing to the raw Wikipedia dump\footnote{\url{https://dumps.wikimedia.org} (enwiki/20210320)}.
For each article, we extract the plain text by WikiExtractor\footnote{\url{https://github.com/attardi/wikiextractor}}.
We split the Wikipedia articles into sentences with the NLTK library\footnote{\url{https://www.nltk.org}} and map entity pairs to candidate relation descriptions with the following steps: 

{\flushleft {\textbf{Entity collection}}}. We collect Wikipedia page titles (surface form) as our entities. To acquire knowledge and utilize the pre-trained entity embeddings in Wikipedia2Vec \cite{yamada2020wikipedia2vec} in the later steps, we only keep entities that can be recognized by Wikipedia2Vec.

{\flushleft {\textbf{Local mention-entity mapping}}}. Wikipedia2Vec uses hyperlinks to collect a global mention-entity dictionary to map the entity mention to the referent entities, like mapping ``apple'' to ``Apple Inc'' or ``Apple (food)''. 
In this work, we follow a similar approach to build the mapping. 
To maintain high accuracy and low ambiguity, we craft the entity mention from the entity by removing the content wrapped by parenthesis and the content after the first comma. For example, a mention-entity pair could be (``Champaign'', ``Champaign, Illinois'') or (``Python'', ``Python (programming language)''). 
Unlike Wikipedia2Vec, we create a local dictionary for each Wikipedia page.
When processing a page, we dynamically update the dictionary with mention-entity pairs collected from the hyperlinks, and extract the entity occurrence with the updating dictionary in one pass. This can reduce the ambiguity when two entities with the same entity mention co-occur on one page and also avoid collecting trivial entity occurrence on the page.

{\flushleft {\textbf{Hyperlink mapping correction}}}.
Using hyperlinks to collect entities will lead to errors under some conditions: 
1) The original link is redirected to a new page, where the title does not match with the entity in the link; 2) The entity in the link is lower-cased and thus, does not match with any title. Under the first condition, we just skip this entity because we require that the entity mention must appear in the sentence to prove its occurrence. Under the second situation, if there is only one page title matching with the entity under the case-insensitive setting, we correct the entity to this page title. Otherwise, if there is more than one match, we use the entity embeddings in Wikipedia2Vec to measure the cosine similarity between each matched title and the title of the current page and correct the entity with the most relevant one.

\begin{table}[tp!]
\small
\setlength\tabcolsep{1.5pt} %
\begin{tabular}{c|p{0.85\linewidth}}
\toprule
\textbf{Rating} & \multicolumn{1}{c}{\textbf{Criterion}} \\
\midrule[0.75pt]
5 & The relation description is explicit, significant, and correct, with which users can understand the relationship correctly and easily. \\
\hline
4 & The relation description is a bit less explicit (reasoning is a bit indirect or description is a bit unclear), less significant (containing a little irrelevant content), and less correct (containing minor errors that do not affect the understanding). \\
\hline
3 & The relation description is fairly explicit, significant, and correct, while users can still understand the relationship. \\
\hline
2 & The relation description is not explicit (reasoning is difficult or description is unclear), significant (containing much irrelevant content), or correct (containing major errors that affect the understanding), while users can still infer the relationship to some extent. \\
\hline
1 & The relation description is completely wrong or does not show any relationship between the two entities. \\
\bottomrule
\end{tabular}
\caption{Annotation guidelines excerpt.}
\vspace{-4mm}
\label{table:rating_scale}
\end{table}

\begin{table*}[ht]
\scriptsize
\setlength\tabcolsep{1.5pt}
\centering
\scalebox{0.93}{
\begin{tabular}{p{0.11\linewidth}|p{0.14\linewidth}|p{0.14\linewidth}|p{0.14\linewidth}|p{0.14\linewidth}|p{0.14\linewidth}|p{0.14\linewidth}}
\toprule
 & \textbf{ExpScore} & \textbf{SigScore} & \textbf{RDScore} & \textbf{RelationSyn-$0$} & \textbf{RelationSyn-$1$} & \textbf{RelationSyn-$5$} \\
\midrule[1pt]
(Mucus, Stomach) & As the first two chemicals may damage the stomach wall, \textit{mucus} is secreted by the \textit{stomach}, providing a slimy layer that acts as a shield against the damaging effects of the chemicals. & The \textit{mucus} produced by these cells is extremely important, as it prevents the \textit{stomach} from digesting itself. & The \textit{mucus} produced by these cells is extremely important, as it prevents the \textit{stomach} from digesting itself. & \textit{Mucus} is a fluid that is produced by the \textit{stomach}. & \textit{Mucus} is the main barrier to mucus from the \textit{stomach}. & \textit{Mucus} is a thick, protective fluid that is secreted by the \textit{stomach}. \\
\hline
(Surfers Paradise, Queensland) & \textit{Surfers Paradise} is a coastal town and suburb in the City of Gold Coast, \textit{Queensland}, Australia. & In 2009 as part of the Q150 celebrations, \textit{Surfers Paradise} was announced as one of the Q150 Icons of \textit{Queensland} for its role as a "location". & \textit{Surfers Paradise} is a coastal town and suburb in the City of Gold Coast, \textit{Queensland}, Australia. & \textit{Surfers Paradise} is a coastal suburb in the City of Brisbane, \textit{Queensland}, Australia. & \textit{Surfers Paradise} is a coastal town and locality in the Shire of Mareeba, \textit{Queensland}, Australia. & \textit{Surfers Paradise} is a coastal suburb in the City of Redland, \textit{Queensland}, Australia. \\
\hline
(Parkinson's disease, Dopamine) & Thus for the first time the reserpine-induced Parkinsonism in laboratory animals and, by implication, \textit{Parkinson's disease} in humans was related to depletion of striatal \textit{dopamine}. & \textit{Parkinson's disease} is characterized by the death of cells that produce \textit{dopamine}, a neurotransmitter. & \textit{Parkinson's disease} is associated with the degeneration of \textit{dopamine} and other neurodegenerative events. & \textit{Parkinson's disease} is a neurodegenerative disease involving the loss of \textit{dopamine} in the brain. & \textit{Parkinson's disease} is a neurodegenerative disease characterized by the loss of \textit{dopamine}. & \textit{Parkinson's disease} is a neurodegenerative disorder characterized by a slow and steady loss of \textit{dopamine} in the substantia nigra. \\
\hline
(Dayan Khan, Oirats) & Mandukhai and \textit{Dayan Khan} defeated the \textit{Oirats} and the taishis who ruled the Yellow River Mongols. & By 1510 \textit{Dayan Khan} had unified the entire Mongol nation including \textit{Oirats}. & By 1510 \textit{Dayan Khan} had unified the entire Mongol nation including \textit{Oirats}. & \textit{Dayan Khan} was a khan of the \textit{Oirats}. & \textit{Dayan Khan} was a khan of the \textit{Oirats}. & \textit{Dayan Khan} defeated the \textit{Oirats} in 1510 with the assistance of the Four Oirats. \\
\hline
(Knowledge, Epistemology) & In \textit{epistemology}, descriptive \textit{knowledge} is knowledge that can be expressed in a declarative sentence or an indicative proposition. & These questions, but particularly the problem of how experience and \textit{knowledge} interrelate, have broad theoretical and practical implications for such academic disciplines as \textit{epistemology}, linguistics, and psychology. & \textit{Knowledge} is the primary subject of the field of \textit{epistemology}, which studies what we know, how we come to know it, and what it means to know something. & In \textit{epistemology}, \textit{knowledge} is a description of the possible meaning of knowledge. & In philosophy, aristocratic \textit{knowledge} is a form of knowledge that can be gained through experience, through the use of a method of \textit{epistemology}. & In the philosophy of \textit{epistemology}, \textit{knowledge} is often referred to as "a priori" or "synthetic". \\
\hline
(Atlantic Coast Line Railroad, Seaboard Air Line Railroad) & The \textit{Atlantic Coast Line Railroad} later merged with the \textit{Seaboard Air Line Railroad} to form the Seaboard Coast Line Railroad. & In 1967, the \textit{Atlantic Coast Line Railroad} merged with the \textit{Seaboard Air Line Railroad}, forming the Seaboard Coast Line Railroad. & In 1967, the \textit{Atlantic Coast Line Railroad} merged with the \textit{Seaboard Air Line Railroad}, forming the Seaboard Coast Line Railroad. & The \textit{Atlantic Coast Line Railroad} was merged into the \textit{Seaboard Air Line Railroad} in 1887. & The \textit{Atlantic Coast Line Railroad} merged with the \textit{Seaboard Air Line Railroad} in 1986 to form CSX Transportation. & The \textit{Atlantic Coast Line Railroad} merged with the \textit{Seaboard Air Line Railroad} on July 1, 1967, to form the Seaboard Coast Line Railroad. \\
\hline
(Twilight, Sunset) & \textit{Twilight} is the period of night after \textit{sunset} or before sunrise when the Sun still illuminates the sky when it is below the horizon. & Near the summer solstice, there are less than 8 hours between \textit{sunset} and sunrise, with \textit{twilight} lasting past 10 pm. & \textit{Twilight} is the period of night after \textit{sunset} or before sunrise when the Sun still illuminates the sky when it is below the horizon. & \textit{Twilight} is the period of daylight between sunrise and \textit{sunset} when the Sun is below the horizon. & \textit{Twilight} is the period of darkness when the Sun is below the horizon. & \textit{Twilight} is the period of darkness from \textit{sunset} to sunrise when the Sun is below the horizon. \\
\hline
(Rock shelter, Cliff) & \textit{Rock shelters} form because a rock stratum such as sandstone that is resistant to erosion and weathering has formed a \textit{cliff} or bluff, ..., and thus undercuts the cliff. & A \textit{rock shelter} is a shallow cave-like opening at the base of a bluff or \textit{cliff}. & A \textit{rock shelter} is a shallow cave-like opening at the base of a bluff or \textit{cliff}. & A \textit{rock shelter} is a structure built on the top of a \textit{cliff}. & A \textit{rock shelter} is a \textit{cliff} or clifftop that is surrounded by a rock. & A \textit{rock shelter} is a small, relatively flat, cave or cave-like structure on a \textit{cliff}. \\
\bottomrule
\end{tabular}
}
\vspace{-1mm}
\caption{Sample of relation descriptions produced by \textit{ExpScore}, \textit{SigScore}, \textit{RDScore}, and \textit{RelationSyn-$m$}.}
\vspace{-4mm}
\label{table:case}
\end{table*}

{\flushleft {\textbf{Filtering}}}.
\label{sec:filtering}
Sometimes the entity mention extracted from the sentence may be part of a bigger noun phrase, which is not an entity mention. For example, suppose we recognize ``algorithm'' and ``graph'' as entity mentions in the sentence ``The breadth-first-search algorithm is a way to explore the vertexes of a graph layer by layer.'' However, this is not a good relation description between``algorithm'' and ``graph'' because the subject is ``breadth-first-search algorithm'' rather than ``algorithm''. Therefore, it is necessary to determine the completed noun phrase for each entity mention. With the dependency tree of the sentence, we recursively find all the child tokens and the ancestor tokens that are connected to the entity mention with a \textit{compound} dependency. To avoid any confusion, we simply reject the entity occurrence if its completed noun phrase and entity mention are different.

Besides, to ensure that the length of relation descriptions is reasonable, we only keep the sentences with the number of tokens $\in [5, 50]$. We also only keep sentences whose shortest dependency path pattern between two target entities starts with \textit{nsubj} or \textit{nsubjpass} (more details are in Section \ref{sec:explicitness}).

\section{Generation Examples}
\label{app:examples}

We present sample outputs of the models in Table~\ref{table:case}, with analysis of the results in Section~\ref{sec:analysis}.

\end{document}